\documentclass{llncs}

\usepackage[T1]{fontenc}
\usepackage{graphicx}
\usepackage{booktabs}
\usepackage{multirow}
\usepackage{amsmath,amssymb}
\usepackage{xspace}
\usepackage{siunitx}
\usepackage{microtype}
\usepackage[hidelinks]{hyperref}
\usepackage{tikz}
\usepackage{listings}
\usepackage{float}
\usepackage{algorithm}
\usepackage{algpseudocode}
\usepackage{bbding}
\usepackage{float}

\graphicspath{{figures/}}
\usetikzlibrary{positioning, fit, arrows.meta, calc, shapes.geometric}

\newcommand{\method}{FuseSampleAgg\xspace}

\newcommand{\codeurl}{\url{https://github.com/SV25-22/FuseSampleAgg}}

\usepackage{enumitem}
\setlist[itemize]{topsep=2pt, itemsep=1pt, parsep=0pt, partopsep=0pt}

\begin{document}

\title{\method: One-Pass Neighborhood Estimation for Budgeted Knowledge-Graph Refresh and Validation}

\author{Aleksandar Stanković\inst{1}\orcidID{0009-0003-5238-7251}\Envelope \and
Haoran Du\inst{2}\orcidID{0009-0000-6542-6940} \and
Xinming Wang\inst{3}\orcidID{0009-0001-9111-4817}}
\institute{University of Novi Sad, Novi Sad, Serbia\\
\email{stankovic.sv25.2022@uns.ac.rs}
\and Shanghai Key Lab of Intelligent Information Processing,\\
College of Computer Science and Artificial Intelligence, Fudan University, China\\
\email{hrdu24@m.fudan.edu.cn}
\and Institute of Automation, Chinese Academy of Sciences, China\\
\email{wangxinming2024@ia.ac.cn}}

\maketitle

\begin{abstract}
Operational knowledge-graph (KG) pipelines in networking and cybersecurity increasingly need to refresh embeddings under strict time, memory, and audit budgets, especially as curated feeds and LLM-assisted extraction accelerate KG updates. A recurring per-step cost in mini-batch KG learning is neighborhood-context estimation: uniform neighbor sampling without replacement followed by mean aggregation. Common frameworks realize this estimator via sampled-subgraph (block) materialization and intermediate feature gathers, adding kernel launches, allocator pressure, and transient memory spikes.
We present \method, a fused PyTorch CUDA operator that samples neighbors and emits the sampled-neighborhood mean directly, avoiding explicit block construction while preserving GraphSAGE-mean semantics for the same sampled neighbor IDs. \method supports seed-controlled sampling and optional saved-index replay for reproducible validation and regression testing. Across large-graph mini-batch workloads, \method improves FP32 end-to-end step latency by 2.24$\times$--3.48$\times$ over tuned DGL baselines and reduces transient GPU memory by up to 160$\times$ in our measurements. On OGB KG completion benchmarks, such as WikiKG2 and BioKG, \method reduces step time and peak VRAM while matching ranking quality within seed variability, improving time-to-quality for budgeted KG refresh.

\end{abstract}

\pagebreak

\section{Introduction}
Knowledge graphs (KGs) increasingly need repeated embedding refresh as curated feeds, schema edits, and LLM-assisted extraction accelerate graph updates in domains such as networking and cybersecurity~\cite{hogan2022knowledgegraphs,wang2025hitchhiker,sikos2023ckg,zhao2024cyberkg,sun2023ctimining,hu2024llmtikg,fieblinger2024actionable}. In these settings, representation learning becomes a recurring \emph{maintenance} task performed under fixed time, GPU-memory, and auditability budgets~\cite{xue2023kgquality,pellegrino2024kgheartbeat}. A persistent per-step bottleneck is neighborhood-context estimation: GraphSAGE-style pipelines sample neighbors, materialize sampled subgraphs into blocks, gather features, and then aggregate, even though explicit block construction is not required by the estimator semantics~\cite{hamilton2017inductive,wang2019dgl,fey2019pyg}.

We present \method, a fused CUDA operator that jointly performs uniform neighbor sampling without replacement and mean aggregation, directly emitting the aggregated context for 1--2 hop encoders while avoiding explicit block materialization. The operator preserves GraphSAGE-mean semantics for fixed sampled neighbor IDs, supports seed-controlled sampling and optional saved-index replay for reproducible validation, and leaves the downstream encoder/decoder unchanged. Across large-graph mini-batch workloads and OGB KG completion benchmarks, \method improves end-to-end step time and reduces transient GPU memory while preserving task metrics within expected stochastic variability.

The paper makes four contributions. First, it frames KG representation learning as a \emph{budgeted refresh} workload where time-to-quality and memory stability are first-order constraints. Second, it formalizes a block-free realization of the sampled-neighborhood mean while preserving GraphSAGE-mean semantics for fixed sampled neighbor IDs. Third, it implements the fused operator as a PyTorch CUDA extension with determinism and optional replay support. Fourth, it evaluates runtime, memory, and task-level behavior on large-graph mini-batch workloads and OGB KG completion benchmarks.

\begin{figure}[H]
    \centering
    \begin{tikzpicture}[
        scale=0.92, every node/.style={transform shape},
        box/.style={draw, rounded corners, align=center, inner sep=3.5pt, font=\scriptsize},
        arr/.style={-Latex, thick},
        grp/.style={draw, rounded corners, dashed, inner sep=6pt, color=gray}
    ]
        \node[box] (sampler) {Neighbor\\Sampling};
        \node[box, right=10mm of sampler] (blocks) {Subgraph / Block\\Materialization};
        \node[box, right=10mm of blocks] (gather) {Feature\\Gather};
        \node[box, right=10mm of gather] (agg) {Aggregation\\(mean)};
        \node[box, right=10mm of agg] (mlp) {MLP / Decoder\\(e.g., DistMult)};

        \draw[arr] (sampler) -- (blocks);
        \draw[arr] (blocks) -- (gather);
        \draw[arr] (gather) -- (agg);
        \draw[arr] (agg) -- (mlp);

        \node[grp, fit=(sampler)(blocks)(gather)(agg), label={[gray]above:{\small Baseline (framework pipeline)}}] (baselinegrp) {};

        \node[box, below=2.5cm of $(blocks.south)!0.5!(gather.south)$] (fsa) {\method\\Fused Sampling +\\Mean Aggregation};
        \node[box, right=15mm of fsa] (mlp2) {MLP / Decoder\\(e.g., DistMult)};

        \draw[arr] (fsa) -- (mlp2);

        \node[grp, fit=(fsa), label={[gray]above:{\small Proposed}}] (fsaGrp) {};

        \draw[arr, red!70!black] (blocks.south) -- ++(0,-0.8) -| node[pos=0.25, left=2pt, align=right, font=\footnotesize, text=black] {eliminate explicit\\block tensors} (fsa.north west);
        \draw[arr, red!70!black] (gather.south) -- ++(0,-0.8) -| node[pos=0.25, right=2pt, align=left, font=\footnotesize, text=black] {avoid intermediate\\feature gathers} (fsa.north east);
    \end{tikzpicture}
    \vspace{-1mm}
    \caption{Estimator realization in budgeted KG refresh. Standard pipelines often realize neighborhood-context estimation by materializing a sampled subgraph and then aggregating. \method computes the same sampled-neighborhood mean estimator \emph{for the same sampled neighbor IDs} directly, avoiding intermediate materialization that is not required by the estimator semantics.}
    \label{fig:pipeline}
\end{figure}

\section{Related Work}
KG representation learning is widely used for completion, validation, and downstream knowledge services, while KG lifecycle work emphasizes continuous monitoring, reproducibility, and governance during maintenance~\cite{hogan2022knowledgegraphs,wang2017kgembedding,xue2023kgquality,pellegrino2024kgheartbeat}. In cybersecurity settings, KGs commonly integrate STIX/TAXII-style threat intelligence and increasingly incorporate LLM-assisted extraction, which further raises refresh frequency~\cite{stix21,taxii2021,mitre_attack_stix,hu2024llmtikg,fieblinger2024actionable}. For scalable training, GraphSAGE-style systems sample bounded neighborhoods and aggregate them, usually through frameworks such as DGL or PyG that materialize sampled blocks before message passing~\cite{hamilton2017inductive,wang2019dgl,fey2019pyg}. Related systems accelerate sampling, loading, or sparse compute through caching, partitioning, optimized loaders, or fused message-passing kernels (e.g., NextDoor, GNNAdvisor, GraphBolt, fuseGNN, TCGNN)~\cite{jangda2021nextdoor,wang2021gnnadvisor,dgl2024graphbolt,chen2021fusegnn,kao2023tcgnn}; orthogonal work changes the sampling policy or estimator itself~\cite{chen2018fastgcn,zou2019ladies,zeng2020graphsaint,chiang2019cluster}.

\paragraph{Where \method Fits.}
\method is complementary to these systems: it preserves the sampling policy and model form, but moves the fusion boundary earlier by computing the sampled-neighborhood mean \emph{without} explicit block construction. This choice also motivates seed-controlled sampling and optional saved-index replay for repeatable validation in maintenance workflows.

\paragraph{Scope Relative to Other Accelerated Graph Frameworks.}
GraphBolt, fuseGNN, and TCGNN are important related systems, but they optimize different parts of the pipeline or assume different execution boundaries. GraphBolt primarily improves sampling and data movement, while fuseGNN and TCGNN focus on fused message passing or sparse compute after graph structure is already available. Our comparison protocol therefore uses tuned end-to-end DGL pipelines as the primary block-materializing baseline for the exact estimator studied here. Accordingly, we do not claim universal dominance over every GPU graph framework; rather, we show that moving the fusion boundary earlier yields consistent gains against a strong tuned production baseline.

\section{Method}

\subsection{Problem Setting and Motivating Application}\label{sec:method_setting}
We study KG representation learning as a recurring \emph{budgeted refresh} operation. As facts, schemas, and extraction rules evolve, representations must be refreshed under fixed time, memory, and auditability constraints. Let $Q(\theta;G)$ denote a downstream validation utility on snapshot $G$ (e.g., validation accuracy or MRR), and define the lifecycle metric
\[
T(\tau)=\inf\{\,t \;:\; Q(\theta_t;G^{(t)}) \ge \tau \,\},
\]
which measures how quickly a refresh run reaches a target validation regime. This viewpoint motivates optimizing the per-step neighborhood-context estimator. A representative application is threat-intelligence KG maintenance, where STIX/TAXII-compatible sources and LLM-assisted extraction accelerate updates and tighten validate$\rightarrow$repair$\rightarrow$retrain loops under shared GPU budgets~\cite{stix21,taxii2021,mitre_attack_stix,mitre_attack_data_tools}. In such settings, the bottleneck is often the sampling--materialization--aggregation path rather than dense tensor math.

\subsection{Fused Sampling and Mean Aggregation}\label{sec:fused}
Let $G=(V,E)$ be stored in CSR $(\texttt{rowptr},\texttt{col})$ with features/embeddings $X\in\mathbb{R}^{|V|\times D}$. Given seed nodes $S$ and fanout $k$, GraphSAGE-mean uses the sampled-neighborhood mean
\[
h_{\mathrm{nei}}(v)=\frac{1}{t}\sum_{u\in U(v)}X[u], \qquad t=\min(k,|N(v)|),
\]
where $U(v)$ is a uniform sample without replacement from $N(v)$ when $|N(v)|\ge k$. Framework baselines typically compute this by materializing sampled blocks and then aggregating.

\textbf{\method.} \method fuses sampling and mean aggregation in one CUDA operator: for each seed $v\in S$, it samples up to $k$ neighbors from CSR and directly accumulates and writes $h_{\mathrm{nei}}(v)$, avoiding explicit block tensors and intermediate gathered-feature tensors. For 2-hop context with fanouts $(k_1,k_2)$, \method computes the nested mean
\[
\hat X_r=\frac{1}{|U_1(r)|}\sum_{u\in U_1(r)}\left(\frac{1}{|W(u)|}\sum_{w\in W(u)}X[w]\right),
\]
skipping empty neighborhoods. For fixed sampled neighbor IDs, \method matches the arithmetic result of block-based aggregation; and under uniform sampling without replacement, the sample mean is an unbiased estimator of the full neighbor mean~\cite{vitter1985reservoir}.

\paragraph{Extension Feasibility Beyond Uniform Mean Aggregation.}
The same early-fusion idea extends directly to weighted averages by streaming weighted numerators and denominators without explicit blocks. Attention-style aggregation is more demanding because it requires online score computation and numerically stable normalization within each sampled neighborhood, and deeper-hop variants increase memory traffic, synchronization, and floating-point error accumulation. We therefore treat relation-aware, attention-based, and deeper-hop operators as natural but unvalidated extensions of the present design.

\pagebreak

\subsection{Semantics and Estimator Properties}\label{sec:theory}
\paragraph{Estimator-Level Semantic Equivalence (Fixed Sampled IDs).}
Fix a CSR graph, a seed set, and sampled neighbor indices. Then \method emits the same arithmetic means as a baseline that (i) materializes the same sampled blocks/subgraph and (ii) applies mean aggregation over that materialization.
\begin{proposition}[Semantic equivalence for fixed sampled indices]\label{prop:equiv}
For fixed sampled neighbor IDs, the neighbor-mean outputs emitted by \method are identical to those obtained by block materialization followed by GraphSAGE-mean aggregation.
\end{proposition}
\begin{proof}[Sketch of proof]
Both realizations compute the same mean over the same feature vectors; materialization changes only the intermediate representation of the sampled neighbor IDs.
\end{proof}

\paragraph{Unbiasedness of the Sampled Neighbor Mean.}
When $|N(v)|\ge k$ and $U(v)$ is a uniformly random $k$-subset of $N(v)$ sampled without replacement, the sample mean is an unbiased estimator of the full neighbor mean.
\begin{lemma}[Unbiased neighbor-mean estimator]\label{lem:unbiased}
Assume $|N(v)|\ge k$ and $U(v)$ is a uniformly random $k$-subset of $N(v)$. Then
\[
\mathbb{E}\!\left[\frac{1}{k}\sum_{u\in U(v)}X[u]\right] = \frac{1}{|N(v)|}\sum_{u\in N(v)}X[u].
\]
\end{lemma}
\begin{proof}[Sketch of proof]
For each $u\in N(v)$, $\Pr[u\in U(v)]=k/|N(v)|$; linearity of expectation gives the result.
\end{proof}

\paragraph{Reproducibility for Validation Workflows.}
\method supports seed-controlled sampling to produce repeatable sampled neighbor IDs for a fixed CSR graph and seed/frontier order. Optionally, \texttt{save\_indices} records sampled IDs in forward and replays them in backward, so gradients correspond to the same sampled neighborhoods. This enables controlled ablations and regression testing in continuous KG maintenance pipelines.

\section{Implementation Summary}
\method is packaged as a PyTorch CUDA extension and operates on contiguous CSR (\texttt{int32}) and contiguous feature tensors. The 1-hop kernel uses a warp-per-seed mapping with shared-memory buffering for sampled indices; the 2-hop kernel uses a block-per-root mapping with small shared buffers for $U$ and $W$. By design, no block tensors are materialized, reducing allocator pressure and intermediate memory traffic.

\paragraph{Complexity.}
For batch size $B$ and feature width $D$: 1-hop requires $\Theta(B\,k\,D)$ feature loads and $\Theta(B\,D)$ writes, while 2-hop requires $\Theta(B\,k_1k_2\,D)$ loads and $\Theta(B\,D)$ writes. In contrast to block-based baselines, \method avoids constructing and traversing intermediate block structures, which drives the observed step-time and memory improvements.

\section{Experimental Setup}
\textbf{Datasets.} We evaluate on Reddit, ogbn-arxiv, and ogbn-products for large-graph mini-batch training, and on ogbl-wikikg2 and ogbl-biokg for KG completion~\cite{hu2020ogb}.

\textbf{Baselines.} \method is compared against DGL pipelines that sample, materialize blocks, and aggregate~\cite{wang2019dgl}; for fairness, we tune three representative DGL modes (GPU graph, CPU workers, CPU+UVA) and report the best per configuration.

\textbf{Precision and metrics.} All experiments use FP32. We report CUDA-synchronized end-to-end step latency (forward+backward+optimizer) together with \emph{peak transient GPU memory} observed during the timed region, excluding persistent model parameters and resident feature storage. KG completion uses a lightweight 2-hop encoder composed with an MLP and DistMult~\cite{yang2015distmult} and reports MRR with the OGB evaluator.

\textbf{Measurement details and fairness.} We precompute a single CSR adjacency for each dataset and reuse it across baselines; unless stated otherwise, we use 20 warmup iterations and time 200 steps, reporting the median latency. Experiments run on a single NVIDIA A800-SXM4-40GB GPU with PyTorch CUDA, DGL, and OGB utilities; scripts that produce raw logs and tables are available at \codeurl.

\section{Results}\label{sec:results}

\subsection{Task-Level Sanity Check: Quality Preservation and Time-to-Quality}
A systems optimization is only useful for KG maintenance if it preserves task quality. In an architecturally matched PyG parity check of the same shallow 2-hop mean encoder, \method shows only small final-accuracy differences that are consistent with sampling stochasticity and floating-point reduction order: on Reddit, test accuracy is 0.9473 versus 0.9491 while wall-clock drops from 3.8\,s to 1.1\,s; on ogbn-products, test accuracy is 0.7114 versus 0.7128 while wall-clock drops from 1.3\,s to 0.9\,s. The practical takeaway is improved \emph{time-to-quality}: \method reaches the same validation regime earlier under the same compute budget.

\begin{table}[H]
    \caption{Accuracy parity summary (final epoch). \textbf{Wall(s)} is the end-to-end wall-clock time to complete the run under the parity setup (same model and hyperparameters), measured on the same system.}
    \label{tab:acc-parity}
    \centering
    \begingroup
    \setlength{\tabcolsep}{3.5pt}
    \renewcommand{\arraystretch}{0.95}
    \small
    \begin{tabular}{llrrrr}
        \toprule
        Dataset & Variant & Train & Val & Test & Wall(s) \\
        \midrule
        ogbn-products & pyg & 0.8848 & 0.8750 & 0.7128 & 1.3 \\
        ogbn-products & fsa & 0.8861 & 0.8738 & 0.7114 & 0.9 \\
        reddit & pyg & 0.9613 & 0.9503 & 0.9491 & 3.8 \\
        reddit & fsa & 0.9614 & 0.9486 & 0.9473 & 1.1 \\
        \bottomrule
    \end{tabular}
    \endgroup

\end{table}

These remaining quality gaps are small and consistent with expected stochasticity from sampling streams and floating-point reduction order rather than a systematic degradation from early fusion. This is important for maintenance-oriented workloads: the systems gain is useful precisely because it does not come with a material task-quality penalty.

\subsection{Budgeted Refresh Efficiency: Runtime and Transient Memory of Neighborhood Estimation}
We next evaluate the practical effect of eliminating explicit subgraph materialization. Each step includes neighborhood estimation, forward/backward, and an optimizer update. For each (dataset, fanout, batch) point, we evaluate three representative DGL modes (\texttt{dgl\_gpu}, \texttt{dgl\_cpu\_workers}, \texttt{dgl\_cpu\_uva}) and compare \method against the \emph{best} DGL mode by median step time, approximating a tuned practitioner baseline.

\begin{table}[H]
    \caption{FP32 end-to-end training-step latency (forward+backward+optimizer, CUDA-synchronized) and peak transient GPU memory during the timed region. Best DGL baseline is selected per configuration among GPU-graph, CPU-workers, and CPU+UVA modes. Speedup is (DGL ms / FSA ms), so values above 1 mean \method is faster.}
    \label{tab:fp32-main-bestdgl}
    \centering
    \begingroup
    \setlength{\tabcolsep}{3.5pt}
    \renewcommand{\arraystretch}{0.95}
    \small
    \resizebox{\linewidth}{!}{
    \begin{tabular}{llrcrrrrr}
        \toprule
        Dataset & Fanout & Batch & Best DGL & DGL ms & FSA ms & Speedup & DGL peak(MB) & FSA peak(MB) \\
        \midrule
        ogbn-arxiv & 15 10 & 512 & dgl\_cpu\_uva & 2.402 & 0.846 & 2.84 & 25.2 & 25.2 \\
        ogbn-arxiv & 15 10 & 1024 & dgl\_cpu\_uva & 2.344 & 0.846 & 2.77 & 27.3 & 25.2 \\
        ogbn-arxiv & 25 10 & 512 & dgl\_gpu & 2.395 & 0.844 & 2.84 & 56.6 & 25.2 \\
        ogbn-arxiv & 25 10 & 1024 & dgl\_gpu & 2.419 & 0.896 & 2.70 & 75.5 & 25.2 \\
        ogbn-products & 15 10 & 512 & dgl\_cpu\_uva & 2.258 & 0.847 & 2.66 & 29.4 & 23.1 \\
        ogbn-products & 15 10 & 1024 & dgl\_cpu\_uva & 2.278 & 0.845 & 2.69 & 33.6 & 25.2 \\
        ogbn-products & 25 10 & 512 & dgl\_cpu\_uva & 2.255 & 0.878 & 2.57 & 31.5 & 23.1 \\
        ogbn-products & 25 10 & 1024 & dgl\_gpu & 2.259 & 0.943 & 2.40 & 4028.6 & 25.2 \\
        reddit & 15 10 & 512 & dgl\_gpu & 3.696 & 1.438 & 2.57 & 3745.5 & 125.8 \\
        reddit & 15 10 & 1024 & dgl\_gpu & 5.438 & 1.563 & 3.48 & 3751.8 & 125.8 \\
        reddit & 25 10 & 512 & dgl\_gpu & 4.630 & 2.066 & 2.24 & 3749.7 & 125.8 \\
        reddit & 25 10 & 1024 & dgl\_gpu & 6.547 & 2.218 & 2.95 & 3756.0 & 125.8 \\
        \bottomrule
    \end{tabular}
    }
    \endgroup

\end{table}

Table~\ref{tab:fp32-main-bestdgl} shows that \method is faster than the best tuned DGL configuration in all reported settings, with FP32 speedups of $2.24\times$--$3.48\times$. When GPU-resident block materialization is fastest, DGL can show multi-GB transient peaks, whereas \method maintains a small and stable footprint by emitting aggregated features directly; when CPU+UVA is strongest, \method still improves latency by removing the remaining materialization and gather overheads. Operationally, lower transient memory and faster steps reduce OOM risk and make fixed-window refresh runs more practical.

\subsection{Knowledge-Graph Completion: Step-Time, Memory, and Time-to-MRR}\label{sec:kg-results}
We also evaluate \method on \texttt{ogbl-wikikg2} and \texttt{ogbl-biokg}~\cite{hu2020ogb} using the same lightweight 2-hop encoder and DistMult decoder across methods; the only difference is whether the 2-hop nested-mean context is computed through block materialization or through the fused operator.

Table~\ref{tab:kg-fp32-lat-mem} reports FP32 training-step latency and peak VRAM (mean$\pm$std over 5 seeds). On BioKG, \method improves step time by $1.44\times$--$1.86\times$ and substantially reduces transient GPU memory (e.g., 738.9\,MB $\rightarrow$ 392.3\,MB at batch 1024), consistent with block construction being a large fraction of per-step cost. On WikiKG2, the speedup is smaller but consistent ($1.15\times$--$1.16\times$) and peak VRAM drops by about 6\%, which is expected because the end-to-end step is more dominated by large embedding tables and KG scoring.
\begin{table}[H]
\caption{KG completion (FP32): training-step latency and peak VRAM (mean$\pm$std over 5 seeds) for a DGL 2-hop baseline that materializes sampled neighborhoods into blocks versus \method which fuses 2-hop sampling and nested-mean aggregation. Speedup is (DGL ms / FSA ms).}

\label{tab:kg-fp32-lat-mem}
\centering
\small
\resizebox{\linewidth}{!}{

\begin{tabular}{llr r r r r r}
\toprule
Dataset & Fanout & Batch & DGL ms & FSA ms & Speedup & DGL peak(MB) & FSA peak(MB) \\
\midrule
ogbl-biokg & 15 10 & 512 & 7.177 ± 0.080 & 3.867 ± 0.052 & 1.86 & 738.7 ± 0.0 & 306.7 ± 0.0 \\
ogbl-biokg & 15 10 & 1024 & 7.764 ± 0.064 & 4.583 ± 0.053 & 1.69 & 738.9 ± 0.0 & 392.3 ± 0.0 \\
ogbl-biokg & 25 10 & 512 & 7.223 ± 0.069 & 4.608 ± 0.067 & 1.57 & 738.7 ± 0.0 & 306.7 ± 0.0 \\
ogbl-biokg & 25 10 & 1024 & 7.780 ± 0.057 & 5.403 ± 0.065 & 1.44 & 738.9 ± 0.0 & 392.3 ± 0.0 \\
ogbl-wikikg2 & 15 10 & 512 & 85.776 ± 0.078 & 73.862 ± 0.036 & 1.16 & 13170.7 ± 0.4 & 12369.2 ± 0.0 \\
ogbl-wikikg2 & 15 10 & 1024 & 86.637 ± 0.099 & 74.647 ± 0.034 & 1.16 & 13193.3 ± 0.4 & 12374.9 ± 0.0 \\
ogbl-wikikg2 & 25 10 & 512 & 85.839 ± 0.046 & 74.464 ± 0.031 & 1.15 & 13181.2 ± 0.6 & 12369.2 ± 0.0 \\
ogbl-wikikg2 & 25 10 & 1024 & 86.745 ± 0.054 & 75.355 ± 0.038 & 1.15 & 13210.3 ± 0.5 & 12374.9 ± 0.0 \\
\bottomrule
\end{tabular}
}

\end{table}

Table~\ref{tab:kg-train6000-mrr} reports WikiKG2 MRR after 6000 training steps (mean$\pm$std over 5 seeds) at batch 1024. \method matches the DGL baseline within run-to-run variability and is slightly higher in our measurements at fanout 25--10. Since WikiKG2 step time drops from 86.704\,ms to 75.960\,ms, 6000 steps finish about 12\% faster; equivalently, within the wall-clock time required by DGL to run 6000 steps, \method can execute about 6850 steps. Thus the systems gain translates directly into improved \emph{time-to-quality} for repeated KG refresh.
\begin{table}[H]
\caption{WikiKG2 final MRR after 6000-step training (mean$\pm$std over 5 seeds).}
\label{tab:kg-train6000-mrr}
\centering
\small
\resizebox{\linewidth}{!}{

\begin{tabular}{llrrlll}
\toprule
Fanout & Variant & Batch & \#Seeds & Step ms (mean$\pm$std) & Valid MRR (mean$\pm$std) & Test MRR (mean$\pm$std) \\
\midrule
15 10 & dgl & 1024 & 5 & 86.619 ± 0.031 & 0.2284 ± 0.0087 & 0.2568 ± 0.0074 \\
15 10 & fsa & 1024 & 5 & 75.097 ± 0.018 & 0.2293 ± 0.0100 & 0.2592 ± 0.0100 \\
25 10 & dgl & 1024 & 5 & 86.704 ± 0.067 & 0.2275 ± 0.0101 & 0.2570 ± 0.0084 \\
25 10 & fsa & 1024 & 5 & 75.960 ± 0.045 & 0.2291 ± 0.0111 & 0.2586 ± 0.0102 \\
\bottomrule
\end{tabular}
}

\end{table}

The KG results also clarify where the method helps most. On BioKG, where the neighborhood estimator is a larger fraction of end-to-end step cost, the gains are larger in both runtime and transient memory. On WikiKG2, the operator still helps, but the step includes heavier embedding-table and scoring work, so the systems improvement is diluted by other dominant costs. This pattern is consistent with the paper's main claim: removing block materialization is most valuable when that path is itself a major bottleneck.

\section{Threats to Validity and Limitations}
Our goal is to accelerate the sampler$\rightarrow$materialize$\rightarrow$aggregate path for a widely used 1--2 hop GraphSAGE-mean primitive; we do not claim state-of-the-art task accuracy or a complete cross-framework leaderboard. Relative performance depends on hardware, runtime behavior, and which baseline mode is strongest for a given platform. Our baselines cover tuned DGL execution modes but not every accelerated graph framework, and our experiments are FP32-only, so the present paper does not establish behavior under FP16/BF16 training. Task metrics remain stochastic because of neighbor sampling, negative sampling, and floating-point reduction order, although \method supports saved-index replay for controlled comparisons. Finally, the KG completion encoder is intentionally lightweight and does not capture relation-aware, attention-based, or deeper-hop message passing; extending early fusion to richer operators is a natural direction for future work.

\section{Conclusion}\label{sec:conclusion}
We presented \method, a block-free realization of the common neighborhood-context estimator formed by uniform neighbor sampling without replacement followed by mean aggregation. By preserving estimator semantics while avoiding intermediate block materialization, \method improves time-to-quality under fixed refresh budgets and supports reproducible validation via seed control and optional replay~\cite{xue2023kgquality,pellegrino2024kgheartbeat}. Across large-graph mini-batch workloads and KG completion benchmarks, \method consistently reduces end-to-end step time and transient GPU memory while preserving task metrics within expected stochastic variability. These results suggest that earlier-fusion systems primitives can make repeated KG refresh and validation substantially more practical. More broadly, they support the design principle that moving the fusion boundary earlier can be valuable even when the downstream model class remains unchanged.

\begin{credits}
\subsubsection{\ackname}
The authors thank the anonymous reviewers for their constructive feedback.

\subsubsection{\discintname}
The authors have no competing interests to declare that are relevant to the content of this article.
\end{credits}

\bibliographystyle{splncs04}
\bibliography{refs}
\end{document}